\documentclass{article}
\usepackage{spconf,amsmath,graphicx}

\usepackage{hyperref}
\usepackage{url}
\usepackage{caption}
\usepackage{subcaption}
\usepackage{comment}
\usepackage{float}
\usepackage{amssymb}
\usepackage{multirow}
\usepackage{hyperref}
\usepackage{xcolor}

\usepackage{enumitem}

\usepackage{algorithm}
\usepackage{algpseudocode}
\usepackage[bottom]{footmisc}

\title{LangVision-LoRA-NAS: Neural Architecture Search for Variable LoRA Rank in Vision--Language Models}

\name{Krishna Teja Chitty-Venkata \qquad Murali Emani \qquad Venkatram Vishwanath}

\address{Argonne National Laboratory, Lemont, Illinois, USA\\
\texttt{\{schittyvenkata, memani, venkat\}@anl.gov}}

\begin{document}
\maketitle
\begin{abstract}
Vision Language Models (VLMs) integrate visual and text modalities to enable multimodal understanding and generation. These models typically combine a Vision Transformer (ViT) as an image encoder and a Large Language Model (LLM) for text generation. LoRA (Low-Rank Adaptation) is an efficient fine-tuning method to adapt pre-trained models to new tasks by introducing low-rank updates to their weights. While LoRA has emerged as a powerful technique for fine-tuning large models by introducing low-rank updates, current implementations assume a fixed rank, potentially limiting flexibility and efficiency across diverse tasks. This paper introduces \textit{LangVision-LoRA-NAS}, a novel framework that integrates Neural Architecture Search (NAS) with LoRA to optimize VLMs for variable-rank adaptation. Our approach leverages NAS to dynamically search for the optimal LoRA rank configuration tailored to specific multimodal tasks, balancing performance and computational efficiency. 
Through extensive experiments using the LLaMA-3.2-11B model on several datasets, LangVision-LoRA-NAS demonstrates notable improvement in model performance while reducing fine-tuning costs. Our Base and searched fine-tuned models on LLaMA-3.2-11B-Vision-Instruct can be found \href{https://huggingface.co/collections/krishnateja95/llama-32-11b-vision-instruct-langvision-lora-nas-6786cac480357a6a6fcc59ee}{\textcolor{blue}{here}} and the code for LangVision-LoRA-NAS can be found \href{https://github.com/krishnateja95/LangVision-NAS}{\textcolor{blue}{here}}. 
\end{abstract}


\section{Introduction} \label{sec:introduction}

\textit{Vision-Language Models} (VLMs) \cite{zhang2024vision} have gained immense popularity across computer vision and NLP tasks, enabling applications such as Image Description and Visual Question Answering (VQA). These models, which integrate vision and language capabilities, are transforming AI interactions beyond what uni-modal models can achieve. Most VLMs use sequential visual representation, converting images into tokens processed alongside language prompts. While this enhances their capabilities, it also introduces considerable computational overhead. Recent VLM models inslude LLaVA \cite{liu2024visual}, Pixtral \cite{agrawal2024pixtral}, Paligemma \cite{beyer2024paligemma}, Qwen-VL \cite{bai2023qwen}, Molmo \cite{deitke2024molmo}.

Low-Rank Adaptation (LoRA) is a parameter-efficient fine-tuning strategy designed to adapt large models to specific tasks with minimal computational overhead. LoRA achieves this by freezing the original model weights and introducing trainable low-rank matrices that approximate weight updates, significantly reducing the number of parameters to fine-tune. For example, in few-shot learning scenarios, LoRA can be effectively applied to a model like CLIP \cite{radford2021learning} to adapt it for image-text matching tasks across multiple datasets. By only fine-tuning the low-rank components, LoRA enables faster training and lower resource consumption. 
Research has shown that LoRA fine-tuned models can outperform fully fine-tuned counterparts in several scenarios, achieving consistent results across datasets and making it an efficient solution for adapting VLMs to new tasks.

Neural Architecture Search (NAS) is a well-studied problem in the computer vision and NLP \cite{chitty2022neural} space, which is defined as a process of automating the design process of neural network architectures to reduce manual trial-and-error methods. It consists of a search space of potential choices and employs strategies such as reinforcement learning or evolutionary algorithms to search for models that maximize accuracy while meeting constraints such as latency. NAS has been instrumental in developing SOTA models for tasks in computer vision and NLP. NAS methods, such as weight-sharing and training-free, have improved the scalability and efficiency of automated methods, enabling their application across different scenarios. Despite its computational demands, NAS continues to push the boundaries of AutoML, offering tailored solutions for diverse datasets and applications.


The current methods for fine-tuning pretrained VLMs (or LLMs) predominantly rely on a uniform LoRA rank across all layers. While this approach simplifies implementation, it imposes significant limitations on the efficacy of LoRA fine-tuning. Specifically, using a low rank (such as 2 or 4) may fail to capture sufficient task-specific information, leading to suboptimal downstream performance, whereas employing a high rank (such as 32 or 64) increases computational overhead, requiring substantial fine-tuning time and storage space. Despite these challenges, the potential benefits of employing LoRA ranks tailored to individual model weights in VLMs or LLMs remain largely unexplored.

To address these limitations, we propose LangVision-LoRA-NAS, a novel and efficient framework that uses neural architecture search (NAS) methodology to identify the optimal LoRA rank for each fully connected (FC) matrix within the model. Unlike conventional approaches, our method introduces flexibility by allowing LoRA ranks to vary across different components of the transformer architecture. Specifically, we define a search space comprising ranks of \{4, 8, 16, 32, 64\} and systematically determine the most suitable rank for each Query (Q) and Value (V) matrix in the transformer layers. This adaptive strategy enables us to balance downstream accuracy with computational efficiency, offering a fast and user-friendly solution for fine-tuning vision-language models on diverse tasks.

The main contributions of our paper are as follows: 
\begin{enumerate}[leftmargin=*, itemsep=1pt, parsep=0pt, topsep=2pt]
\item We propose LangVision-LoRA-NAS, a weight-sharing NAS method to identify the optimal LoRA rank in vision language models. Unlike existing approaches that rely on uniform LoRA ranks, our method enables layer-wise rank customization, improving the balance between downstream performance and computational efficiency.
\item We demonstrate the effectiveness of our methodology on the open-source vision-language model, LLaMA-3.2-11B-Vision. Our experiments showcase that LangVision-LoRA-NAS can efficiently search for and determine optimal LoRA ranks, leading to improved fine-tuning performance while reducing computational and storage overhead. For instance, our method achieves a 2.6x 
reduction in the number of LoRA parameters for LLaMA-3.2-11B model with  DoCVQA dataset from 268.7M (uniform rank 64) to 103.3M (mixed-rank model).
\end{enumerate}


\section{Background and Experimental Setup} \label{sec:background}


\subsection{Attention Mechanism}

The input tensor to LLMs is represented by $\mathbf{X} \in \mathbb{R}^{n \times d}$, where $n$ and $d$ are sequence length and hidden dimension, respectively. The tensor $\mathbf{X}$ is passed through multiple transformer blocks, each module consisting of multi-head self-attention and feed-forward neural network. The Query ($\mathbf{Q}$), Key ($\mathbf{K}$), and Value ($\mathbf{V}$) 
projections are obtained by passing the input $\mathbf{X}$ to three different learned FC layers as follows: $\mathbf{Q} = \mathbf{X} \mathbf{W}_Q, \quad \mathbf{K} = \mathbf{X} \mathbf{W}_K, \quad \mathbf{V} = \mathbf{X} \mathbf{W}_V$
The attention is calculated as $\text{softmax}\left(\frac{\mathbf{Q} \mathbf{K}^\top}{\sqrt{d_k}}\right) \mathbf{V}$.







\begin{figure}
    \centering
     \includegraphics[width=0.7\linewidth]{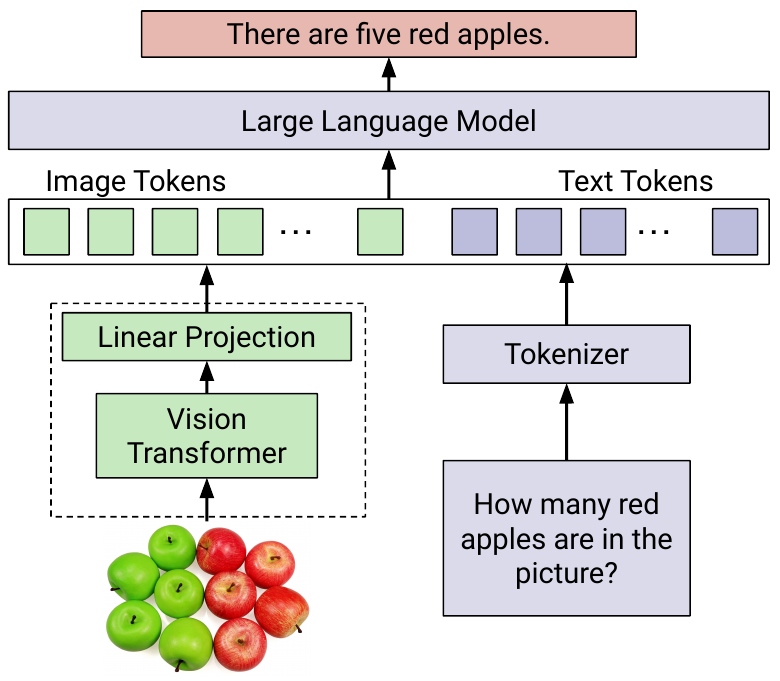}
        \caption{ViT-LLM in Vision Language Model (VLM)}
        \label{fig:VLM}
    \captionsetup{justification=centering}
\end{figure}

\subsection{Vision Language Models (VLMs)} 

Large Language Models (LLMs) are built on transformer architecture \cite{vaswani2017attention}, leveraging the self-attention mechanism to generate human-like text. VLMs integrate visual and textual modalities through three core components: an Image Encoder, a Text Encoder, and a Fusion mechanism. The image encoder, based on Vision Transformers, extracts high-dimensional visual features, while the text encoder, typically an LLM, generates text output, while the fusion layer combines different modalities. These architectures are optimized for multimodal tasks such as image captioning and visual question answering through pretraining objectives like masked token prediction and cross-modal matching.


\vspace{-0.8ex}

\subsection{VLM LoRA Finetuning}

\begin{figure}[H]
    \centering
     \includegraphics[width=\linewidth]{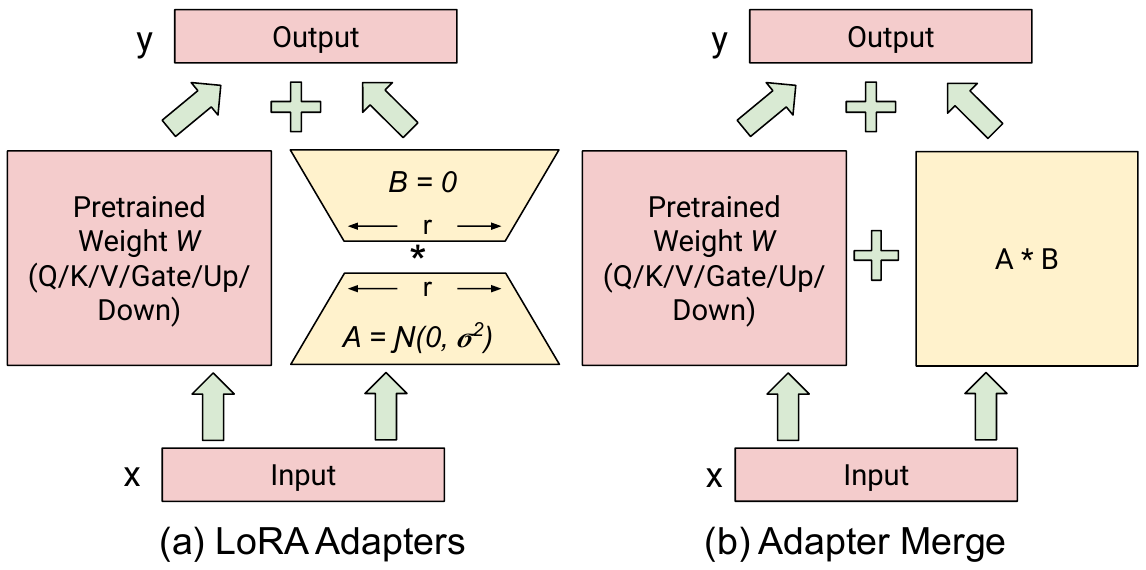}
        \caption{LoRA}
        \label{fig:LoRA}
    \captionsetup{justification=centering}
\end{figure}

LoRA fine-tuning for VLMs introduces trainable low-rank weights to approximate weight updates while keeping the pretrained model weights frozen. Specifically, the weight update $\Delta$W decomposed into the product of two smaller matrices A (m$\times$r) and B (r$\times$n), where r $<<$ m and n, as shown in Figure \ref{fig:LoRA}(a). This decomposition reduces the number of trainable parameters from m$\times$n to m$\times$r + r$\times$n, enabling efficient adaptation with less than 1\% of the original model's parameters being updated. In VLMs, LoRA targets specific weight matrices, such as q$_{proj}$ and v$_{proj}$, to align visual and textual embeddings effectively. The low-rank matrices are merged with the original weights during inference, resulting in no additional latency, illustrated in Figure \ref{fig:LoRA}(b). LoRA allows VLMs to adapt both vision and text encoders independently or jointly, making it highly effective for various tasks.

\subsection{Models and Datasets}

We consider LLaMA-3.2-11B-Vision-Instruct model \cite{Llama_11B_Vision_Instruct}, which combines ViT and an LLM. The vision encoder uses a 32-layer transformer for local image processing, followed by an 8-layer global transformer encoder with gated attention. The visual features are fed into a 40-layer language model via cross-attention layers. 
We use the following datasets as downstream applications from Cauldron \cite{laurençon2024matters}: 
ai2d \cite{kembhavi2016diagram}, 
chartqa \cite{masry2022chartqa},
docvqa \cite{mathew2021docvqa}, 
infographic vqa \cite{mathew2022infographicvqa}, 
intergps \cite{lu2021inter}, 
tqa \cite{kembhavi2017you}, 
vsr \cite{liu2023visual},
vqarad \cite{lau2018dataset},
hitab \cite{cheng2021hitab},
geomverse \cite{kazemi2023geomverse}. 
\raggedbottom

\section{LangVision-LoRA-NAS}


\subsection{Search Space}
Our search space is comprised of three different combinations of Query (Q), Key (K), Value (V), Out (O), Gate (G), Up (U), Down (D), FC1 and FC2 matrices in the self-attention and MLP layers with rank sizes of \{4, 8, 16, 32, 64\}. 
Lower ranks (e.g., 4 or 8) emphasize computational efficiency and reduced parameter counts, which are beneficial for resource-constrained environments. Higher ranks (e.g., 32 or 64) allow the model to capture richer representations at the expense of increased computational overhead. 


\vspace{-1.5ex}

\subsection{Single Path NAS}
\begin{figure}
    \centering
     \includegraphics[width=\linewidth]{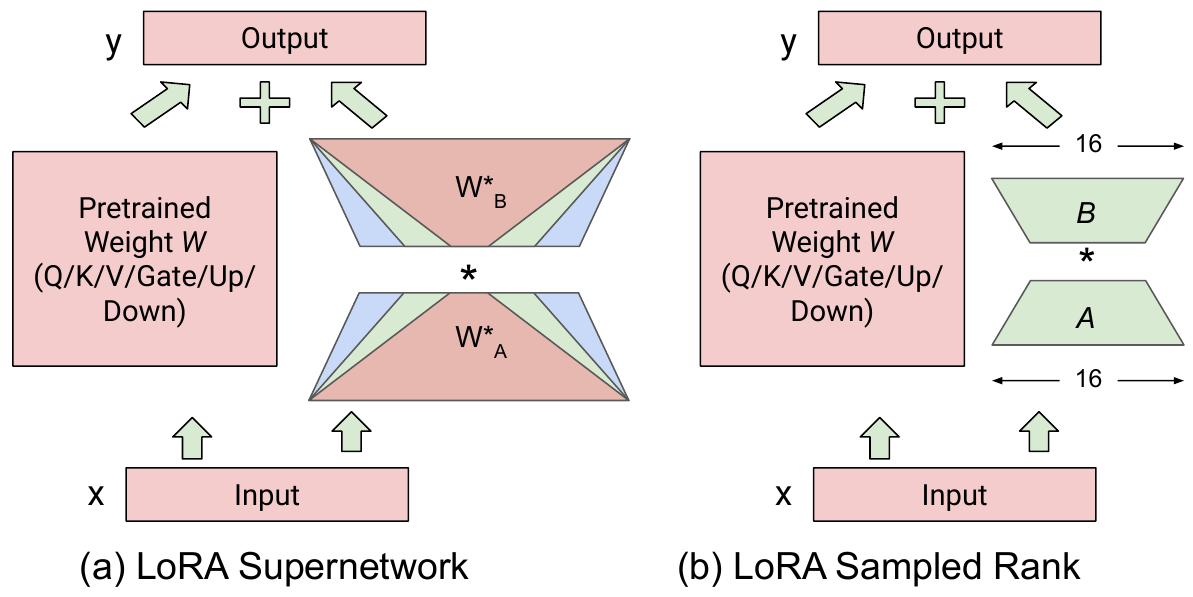}
        \caption{LoRA Supernetwork. The LoRA rank search space consists of 8, 16 and 32 and the rank in Supernetwork is 32.}
        \label{fig:SP_NAS_1}
    \captionsetup{justification=centering}
\end{figure}

\begin{figure}
    \centering
     \includegraphics[width=\linewidth]{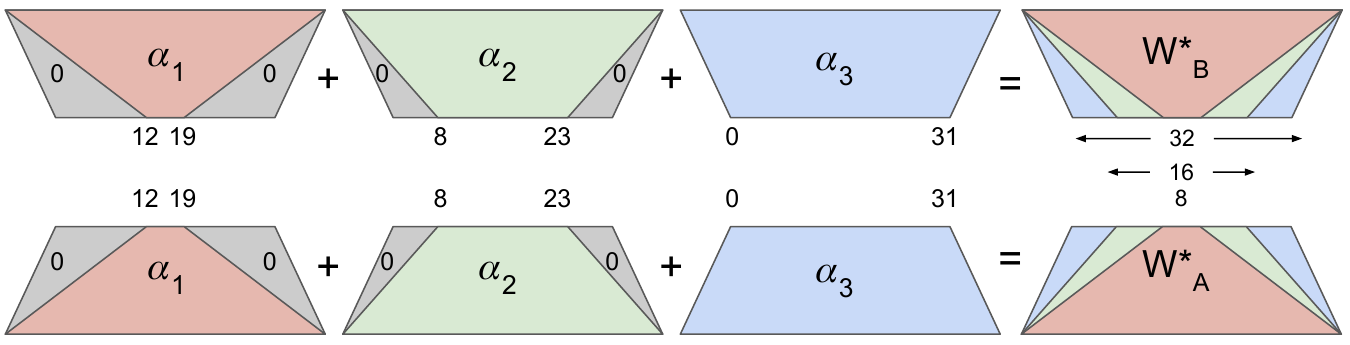}
        \caption{LoRA Superweight}
        \label{fig:SP_NAS_2}
    \captionsetup{justification=centering}
\end{figure}

We propose a one-shot gradient-based search method \cite{liu2018darts, stamoulis2019single} that leverages weight sharing to efficiently determine optimal LoRA ranks. We construct a Supernetwork that incorporates all possible LoRA ranks within the predefined search space. This Supernetwork is trained only once during the fine-tuning phase, significantly reducing computational overhead. The optimal rank configuration is then selected by sampling from the fine-tuned LoRA Supernetwork. The Supernetwork is initialized following standard LoRA procedures, but with the rank set to the highest value in the search space. For example, if the search space is \{8, 16, 32\}, the Supernetwork is initialized with rank 32, as illustrated in Figure \ref{fig:SP_NAS_1}(a) and \ref{fig:SP_NAS_2}. We initialize a set of architectural weight parameters ($\alpha$), one for each discrete LoRA rank in the search space. During the forward pass of fine-tuning, we first compute a softmax of these architectural weights (Equation \ref{eq:softmax}), ensuring that the resulting probabilities lie within the range of 0 and 1.

\begin{equation} \label{eq:softmax}
\alpha_{i} = \frac{\exp(\alpha_{i})}{\sum_{n=1}^{N} \exp(\alpha_{n})} = \frac{e^{\alpha_{i}}}{\sum_{n=1}^{N} e^{\alpha_{n}}}
\end{equation}

Using these weights, we derive the Superweights $\textit{W*}_{A}$ and $\textit{W*}_{B}$, represented as an $\alpha$-weighted sum of all subweight tensors. This process is described in Equations \ref{eq:Wa} and \ref{eq:Wb}.
\begin{equation} \label{eq:Wa}
\textit{W*}_{A} = \alpha_1 \cdot W_{A}[:,12{:}20] + \alpha_2 \cdot W_{A}[:,8{:}24] + \alpha_3 \cdot W_{A}
\end{equation}
\begin{equation} \label{eq:Wb}
\textit{W*}_{B} = \alpha_1 \cdot W_{B}[12{:}20,:] + \alpha_2 \cdot W_{B}[8{:}24,:] + \alpha_3 \cdot W_{B}
\end{equation}
Here, W$_{A}$[:,12{:}20], W$_{A}$[:,8{:}24] and W$_{A}$ represent the weight slices of A weight of LoRA ranks 8, 16 and 32. Each slice is multiplied with the corresponding architectural weight ($\alpha$), and the resulting tensors are summed to construct $\textit{W*}_{B}$ and $\textit{W*}_{B}$. This approach ensures that only one A and B weight exists for forward and backward passes. The input dimensions for LoRA matrix A and output dimensions for LoRA matrix B remain consistent in the Supernetwork. It is essential to use the same set of architectural parameters ($\alpha$) for both LoRA A and B matrices. This is to ensure that their ranks are identical for both trainable weights. There exists only one linear operation in the Supernetwork, irrespective of the number of LoRA rank choices in the search space.

\subsection{LoRA Supernetwork Finetuning}
The LoRA fine-tuning process with the Supernetwork involves two key stages: (a) training the Supernetwork and selecting the optimal LoRA rank based on architectural parameters ($\alpha$), and (b)  end-to-end LoRA fine-tuning with the chosen rank. This approach is efficient, as the Supernetwork requires training only once for a few epochs, unlike evolutionary search methods that repeatedly train and evaluate sampled networks. Initially, all $\alpha$ values are uniformly initialized to give equal importance to all ranks. Training alternates between two steps within a single mini-batch as follows: (1) freeze $\alpha$ and update LoRA weights (A and B) on the training dataset, (2) freeze A and B weights and update $\alpha$ on the validation dataset. The differentiable Superweight formulation allows traditional gradient descent for LoRA rank search. After Supernetwork LoRA finetuning for the given search epochs, the final mixed-rank LoRA model is derived by selecting the rank with the highest $\alpha$ of the target module. 
For example, if the search space is \{8, 16, 32\}, as shown in Figure \ref{fig:SP_NAS_1}(a) and if $\alpha_{2}$ = max($\alpha_{1}$, $\alpha_{2}$, $\alpha_{3}$), then the sampled rank for the layer is 16, as per Figure \ref{fig:SP_NAS_1}(b). 
The pseudo-code of our method is given in Algorithm \ref{Alg:LangVision-LoRA-NAS}.

\begin{algorithm}
\caption{LangVision-LoRA-NAS}
\label{Alg:LangVision-LoRA-NAS}
\begin{algorithmic}[1]
\State \textbf{Input} Pretrained LLM/VLM Model Weights W, LoRA Rank Search Space \{8, 16, 32, 64\}, Target LoRA Modules (Q, K, V, O, G, U, D, FC1, FC2)
\State \textbf{Initialize} LoRA Weights A and B, and NAS trainable architectural parameters {$\alpha_{1}, \alpha_{2}, ...\alpha_{n}$}
\For{search epochs}
    \State Update the LoRA Weight parameters A and B by descending $\nabla_{w}$ $\mathcal{L}_{\text{train}}$(W, A, B, $\alpha$) while freezing $\alpha$.
    \State Update the architectural parameters $\alpha$ by descending $\nabla_{\alpha}$ $\mathcal{L}_{\text{val}}$(W, A, B, $\alpha$) by freezing A and B.
\EndFor
\State Sample the best rank r* based on max({$\alpha_{1}, \alpha_{2}, ...\alpha_{n}$}).
\end{algorithmic}
\end{algorithm}

\vspace{-5mm}

\section{Results and Discussion}

\begin{table*}[]
\centering
\small
\caption{Performance Metrics (Evaluation Perplexity, LoRA Parameter Count and Per epoch finetuning time (excluding search time)) of LLaMA-3.2-11B-Vision-Instruct. LoRA adapters in the transformer model include Query (Q), Key (K), Value (V), Out (O), Gate (G), Up (U), Down (D), FC1 and FC2 Weight Matrices. The base model is based on a LoRA rank of 64, while the \textit{searched} models derived from our  LangVision-LoRA-NAS technique are searched on the following search space: 8, 16, 32, and 64. All models are finetuned for 10 epochs}. 
\label{table:results}
\begin{tabular}{|c|c|c|c|c|c|c|c|c|c|}
\hline
Dataset & \begin{tabular}[c]{@{}c@{}}LoRA\\ Adapters\end{tabular} & Model & \begin{tabular}[c]{@{}c@{}}Eval\\ Per.\end{tabular}     & \begin{tabular}[c]{@{}c@{}}\# LoRA \\ Params\end{tabular}             & \begin{tabular}[c]{@{}c@{}}Epoch Time\\ (Seconds)\end{tabular} & Dataset                                                               & \begin{tabular}[c]{@{}c@{}}Eval\\ Per.\end{tabular}     & \begin{tabular}[c]{@{}c@{}}\# LoRA \\ Params\end{tabular}             & \begin{tabular}[c]{@{}c@{}}Epoch Time\\ (Seconds)\end{tabular} \\ \hline

    \multirow{3}{*}{docvqa} & \begin{tabular}[c]{@{}c@{}}Q,K,V,O\\ G, U, D, \\ FC1, FC2\end{tabular} & 
    \begin{tabular}[c]{@{}c@{}}Base\\ Searched\end{tabular} & 
    \begin{tabular}[c]{@{}c@{}}1.154\\ 1.1539\end{tabular} & 
    \begin{tabular}[c]{@{}c@{}}268.7M (2.5\%)\\ 103.3M (1.0\%)\end{tabular} & 
    \begin{tabular}[c]{@{}c@{}}1815.3\\ 1786.2\end{tabular}&

    \multirow{3}{*}{\begin{tabular}[c]{@{}c@{}}Infographic\\ VQA\end{tabular}} & 
    \begin{tabular}[c]{@{}c@{}}1.416\\ 1.4158\end{tabular} & 
    \begin{tabular}[c]{@{}c@{}}268.7M (2.5\%)\\ 103.3M (1.0\%)\end{tabular} &
    \begin{tabular}[c]{@{}c@{}}463.2\\ 457.8\end{tabular} \\ \cline{2-6} \cline{8-10} & G, U, D & 
    \begin{tabular}[c]{@{}c@{}}Base\\ Searched\end{tabular} & 

    \begin{tabular}[c]{@{}c@{}}1.154\\ 1.1545\end{tabular}  & 
    \begin{tabular}[c]{@{}c@{}}141.6M (1.3\%)\\ 61.6M (0.6\%)\end{tabular} &
    \begin{tabular}[c]{@{}c@{}}1317.2\\ 1310.3\end{tabular} & &

    \begin{tabular}[c]{@{}c@{}}1.417\\ 1.4166\end{tabular}  & 
    \begin{tabular}[c]{@{}c@{}}141.6M (1.3\%)\\ 61.6M (0.6\%)\end{tabular} & 
    \begin{tabular}[c]{@{}c@{}}362.6\\ 362.8\end{tabular} \\ \cline{2-6} \cline{8-10} & Q, K & 

    \begin{tabular}[c]{@{}c@{}}Base\\ Searched\end{tabular} & 

    \begin{tabular}[c]{@{}c@{}}1.166\\ 1.1661\end{tabular} & 
    \begin{tabular}[c]{@{}c@{}}47.2M (0.4\%)\\ 18.0M (0.2\%)\end{tabular} & 
    \begin{tabular}[c]{@{}c@{}}1359.7\\ 1342.8\end{tabular} & & 

    \begin{tabular}[c]{@{}c@{}}1.506\\ 1.5057\end{tabular} & 
    \begin{tabular}[c]{@{}c@{}}47.2M (0.4\%)\\ 18.0M (0.2\%)\end{tabular} & 
    \begin{tabular}[c]{@{}c@{}}373.6\\ 371.6\end{tabular} 
    \\ \hline

    \multirow{3}{*}{vsr} & \begin{tabular}[c]{@{}c@{}}Q,K,V,O \\ G, U, D, \\ FC1, FC2\end{tabular} & 
    \begin{tabular}[c]{@{}c@{}}Base\\ Searched\end{tabular} & 
    \begin{tabular}[c]{@{}c@{}}1.135\\ 1.1351\end{tabular} & 
    \begin{tabular}[c]{@{}c@{}}268.7M (2.5\%)\\ 103.3M (1.0\%)\end{tabular} & 
    \begin{tabular}[c]{@{}c@{}}396.0\\ 391.3\end{tabular}&

    \multirow{3}{*}{intergps} & 
    \begin{tabular}[c]{@{}c@{}}1.314\\ 1.3138\end{tabular} & 
    \begin{tabular}[c]{@{}c@{}}268.7M (2.5\%)\\ 103.3M (1.0\%)\end{tabular} &
    \begin{tabular}[c]{@{}c@{}}275.4\\ 271.1\end{tabular} \\ \cline{2-6} \cline{8-10} & G, U, D & 
    \begin{tabular}[c]{@{}c@{}}Base\\ Searched\end{tabular} & 

    \begin{tabular}[c]{@{}c@{}}1.145\\ 1.1448\end{tabular}  & 
    \begin{tabular}[c]{@{}c@{}}141.6M (1.3\%)\\ 61.6M (0.6\%)\end{tabular} &
    \begin{tabular}[c]{@{}c@{}}285.0\\ 284.7\end{tabular} & &

    \begin{tabular}[c]{@{}c@{}}1.318\\ 1.3184\end{tabular}  & 
    \begin{tabular}[c]{@{}c@{}}141.6M (1.3\%)\\ 61.6M (0.6\%)\end{tabular} & 
    \begin{tabular}[c]{@{}c@{}}211.3\\ 208.4\end{tabular} \\ \cline{2-6} \cline{8-10} & Q, K & 

    \begin{tabular}[c]{@{}c@{}}Base\\ Searched\end{tabular} & 

    \begin{tabular}[c]{@{}c@{}}1.165\\ 1.165\end{tabular} & 
    \begin{tabular}[c]{@{}c@{}}47.2M (0.4\%)\\ 18.0M (0.2\%)\end{tabular} & 
    \begin{tabular}[c]{@{}c@{}}288.3\\ 285.6\end{tabular} & & 

    \begin{tabular}[c]{@{}c@{}}1.37\\ 1.3702\end{tabular} & 
    \begin{tabular}[c]{@{}c@{}}47.2M (0.4\%)\\ 18.0M (0.2\%)\end{tabular} & 
    \begin{tabular}[c]{@{}c@{}}214.9\\ 210.2\end{tabular} 
    \\ \hline

    \multirow{3}{*}{vqarad} & \begin{tabular}[c]{@{}c@{}}Q,K,V,O \\ G, U, D, \\ FC1, FC2\end{tabular} & 
    \begin{tabular}[c]{@{}c@{}}Base\\ Searched\end{tabular} & 
    \begin{tabular}[c]{@{}c@{}}1.719\\ 1.7173\end{tabular} & 
    \begin{tabular}[c]{@{}c@{}}268.7M (2.5\%)\\ 103.3M (1.0\%)\end{tabular} & 
    \begin{tabular}[c]{@{}c@{}}72.9\\ 69.7\end{tabular}&

    \multirow{3}{*}{hitab} & 
    \begin{tabular}[c]{@{}c@{}}1.425\\ 1.4259\end{tabular} & 
    \begin{tabular}[c]{@{}c@{}}268.7M (2.5\%)\\ 103.3M (1.0\%)\end{tabular} &
    \begin{tabular}[c]{@{}c@{}}942.3\\ 938.6\end{tabular} \\ \cline{2-6} \cline{8-10} & G, U, D & 
    \begin{tabular}[c]{@{}c@{}}Base\\ Searched\end{tabular} & 

    \begin{tabular}[c]{@{}c@{}}1.789\\ 1.7865\end{tabular}  & 
    \begin{tabular}[c]{@{}c@{}}141.6M (1.3\%)\\ 61.6M (0.6\%)\end{tabular} &
    \begin{tabular}[c]{@{}c@{}}56.8\\ 54.9\end{tabular} & &

    \begin{tabular}[c]{@{}c@{}}1.446\\ 1.4458\end{tabular}  & 
    \begin{tabular}[c]{@{}c@{}}141.6M (1.3\%)\\ 61.6M (0.6\%)\end{tabular} & 
    \begin{tabular}[c]{@{}c@{}}798.7\\ 797.2\end{tabular} \\ \cline{2-6} \cline{8-10} & Q, K & 

    \begin{tabular}[c]{@{}c@{}}Base\\ Searched\end{tabular} & 

    \begin{tabular}[c]{@{}c@{}}2.235\\ 2.2332\end{tabular} & 
    \begin{tabular}[c]{@{}c@{}}47.2M (0.4\%)\\ 18.0M (0.2\%)\end{tabular} & 
    \begin{tabular}[c]{@{}c@{}}56.4\\ 55.2\end{tabular} & & 

    \begin{tabular}[c]{@{}c@{}}1.95\\ 1.9453\end{tabular} & 
    \begin{tabular}[c]{@{}c@{}}47.2M (0.4\%)\\ 18.0M (0.2\%)\end{tabular} & 
    \begin{tabular}[c]{@{}c@{}}804.7\\ 800.8\end{tabular} 
    \\ \hline

    \multirow{3}{*}{tqa} & \begin{tabular}[c]{@{}c@{}}Q,K,V,O \\ G, U, D, \\ FC1, FC2\end{tabular} & 
    \begin{tabular}[c]{@{}c@{}}Base\\ Searched\end{tabular} & 
    \begin{tabular}[c]{@{}c@{}}1.14\\ 1.1401\end{tabular} & 
    \begin{tabular}[c]{@{}c@{}}268.7M (2.5\%)\\ 103.3M (1.0\%)\end{tabular} & 
    \begin{tabular}[c]{@{}c@{}}317.4\\ 315.0\end{tabular}&

    \multirow{3}{*}{geomverse} & 
    \begin{tabular}[c]{@{}c@{}}1.022\\ 1.0217\end{tabular} & 
    \begin{tabular}[c]{@{}c@{}}268.7M (2.5\%)\\ 103.3M (1.0\%)\end{tabular} &
    \begin{tabular}[c]{@{}c@{}}1923.3\\ 1908.8\end{tabular} \\ \cline{2-6} \cline{8-10} & G, U, D & 
    \begin{tabular}[c]{@{}c@{}}Base\\ Searched\end{tabular} & 

    \begin{tabular}[c]{@{}c@{}}1.143\\ 1.1433\end{tabular}  & 
    \begin{tabular}[c]{@{}c@{}}141.6M (1.3\%)\\ 61.6M (0.6\%)\end{tabular} &
    \begin{tabular}[c]{@{}c@{}}250.0\\ 249.7\end{tabular} & &

    \begin{tabular}[c]{@{}c@{}}1.023\\ 1.0231\end{tabular}  & 
    \begin{tabular}[c]{@{}c@{}}141.6M (1.3\%)\\ 61.6M (0.6\%)\end{tabular} & 
    \begin{tabular}[c]{@{}c@{}}1565.6\\ 1567.4\end{tabular} \\ \cline{2-6} \cline{8-10} & Q, K & 

    \begin{tabular}[c]{@{}c@{}}Base\\ Searched\end{tabular} & 

    \begin{tabular}[c]{@{}c@{}}1.172\\ 1.1721\end{tabular} & 
    \begin{tabular}[c]{@{}c@{}}47.2M (0.4\%)\\ 18.0M (0.2\%)\end{tabular} & 
    \begin{tabular}[c]{@{}c@{}}258.5\\ 255.8\end{tabular} & & 

    \begin{tabular}[c]{@{}c@{}}1.175\\ 1.1747\end{tabular} & 
    \begin{tabular}[c]{@{}c@{}}47.2M (0.4\%)\\ 18.0M (0.2\%)\end{tabular} & 
    \begin{tabular}[c]{@{}c@{}}1597.9\\ 1594.8\end{tabular} 
    \\ \hline

    \multirow{2}{*}{ai2d} & \begin{tabular}[c]{@{}c@{}}Q, K, V, O,\\ G, U, D, \\ FC1, FC2\end{tabular} & \begin{tabular}[c]{@{}c@{}}Base\\ Searched\end{tabular} & \begin{tabular}[c]{@{}c@{}}1.0574\\ 1.0575\end{tabular}   & \begin{tabular}[c]{@{}c@{}}268.7M (2.5\%)\\ 103.3M (1.0\%)\end{tabular} & \begin{tabular}[c]{@{}c@{}}581.33\\ 575.463\end{tabular}       & \multirow{2}{*}{ChartQA} & \begin{tabular}[c]{@{}c@{}}1.3336\\ 1.334\end{tabular}    & \begin{tabular}[c]{@{}c@{}}268.7M (2.5\%)\\ 103.3M (1.0\%)\end{tabular} & \begin{tabular}[c]{@{}c@{}}2896\\ 2847\end{tabular}            \\ \cline{2-6} \cline{8-10} 
                      & G, U, D                                                                    & \begin{tabular}[c]{@{}c@{}}Base\\ Searched\end{tabular} & \begin{tabular}[c]{@{}c@{}}1.0591\\ 1.0591\end{tabular}   & \begin{tabular}[c]{@{}c@{}}141.6M (1.3\%)\\ 61.6M (0.6\%)\end{tabular}  & \begin{tabular}[c]{@{}c@{}}486.24\\ 484.34\end{tabular}        &                          & \begin{tabular}[c]{@{}c@{}}1.3336\\ 1.336\end{tabular}    & \begin{tabular}[c]{@{}c@{}}141.6M (1.3\%)\\ 61.6M (0.6\%)\end{tabular}  & \begin{tabular}[c]{@{}c@{}}1929\\ 1928\end{tabular}            \\ \hline

    \end{tabular}
\end{table*}







\begin{figure*}
    \centering
     \includegraphics[width=0.9\linewidth]{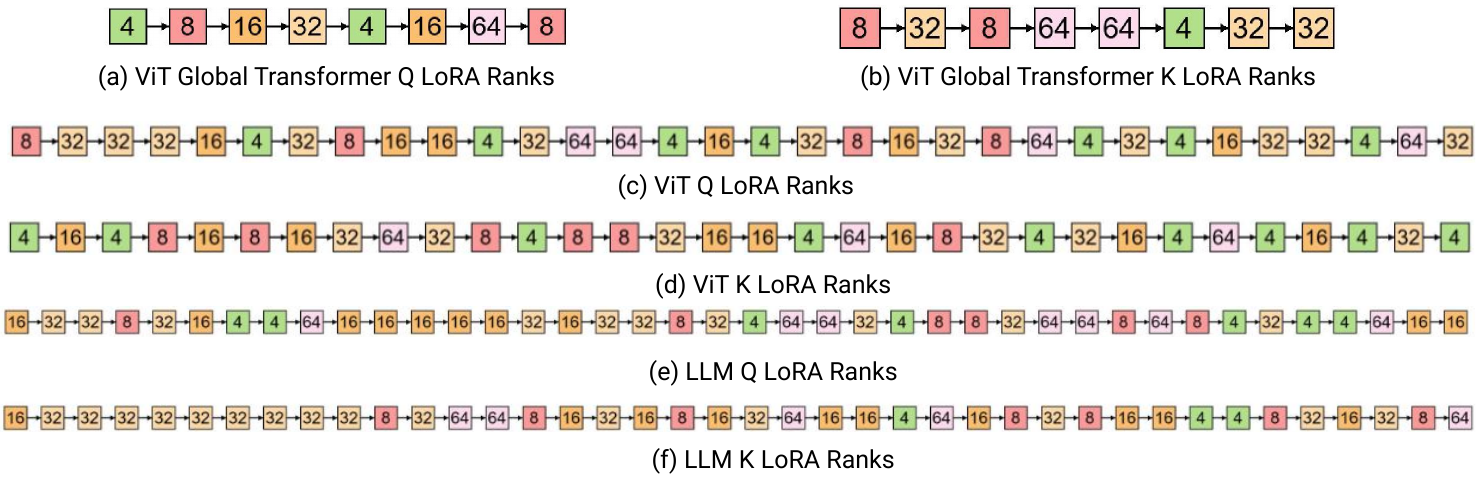}
        \caption{ViT-LLM Searched LoRA Ranks for LLaMA-3.2-11B-Vision-Instruct model on DocVQA Dataset. The target modules are Q and K in both ViT and LLM modules. The LoRA Rank Search Space is \{4, 8, 16, 32, 64\}} 
        \label{fig:docvqa_Q_FC_Searched_model}
    \captionsetup{justification=centering}
\end{figure*}

We evaluate the performance of our LangVision-LoRA-NAS method by analyzing the eval perplexity, LoRA adapter memory footprint and finetuning epoch time of the searched model and the baseline model (uniform rank of 64). We finetune both models using the same batch size, learning rate and hardware platform. Figure \ref{fig:docvqa_Q_FC_Searched_model} illustrates an example of searched ranks for the LLaMA-3.2-11B-Vision-Instruct model with Q and K as target modules and LoRA rank search space of \{4, 8, 16, 32, 64\}. The searched model has less number of LoRA parameters than the base model and rank 16 emerges as a consistent sweet spot for balancing perplexity and \#parameters.


\raggedbottom



Table \ref{table:results} highlights key metrics such as perplexity, LoRA parameter count and average per epoch finetuning time when comparing the uniform-rank LoRA baseline model to the optimized searched mixed-rank model for different datasets. The results demonstrate that the searched models consistently achieve comparable or better performance than their base counterparts while significantly reducing the number of LoRA parameters. For example, we observed a 2.6x reduction in number of parameters on the searched LoRA ranks on the DocVQA dataset. Despite the reduced parameter count, the perplexity of the searched models remained nearly identical or slightly better than the baseline models, suggesting our optimization method does not sacrifice model quality for computational gains. The epoch time reductions observed in searched models also highlight an improvement in efficiency. For instance, the searched ranks on the DocVQA dataset with full adapters (Q,K,V,O,G,U,D,FC1,FC2), the average epoch time decreased from 1815.3s to 1786.2s (~1.6\% improvement). The datasets with higher complexity, such as GeomVerse and InfographicVQA, still benefited from the optimization, indicating the robustness of the approach across varying domains. 
The consistent computational performance across diverse datasets suggests the general applicability of the optimization technique. 

The key takeaways of our paper are as follows: 

\textbf{Efficient LoRA Adapter Rank Search:} Our method identifies LoRA ranks with minimal parameter counts while matching the baseline perplexity across diverse datasets to balance model efficiency and performance. 



\textbf{Enhanced Epoch Time Performance:} The searched LoRA configurations demonstrate faster epoch times. This becomes increasingly pronounced in the case of high fine-tuning epochs and dataset size. This scalability is particularly advantageous for large-scale training tasks.

\textbf{Multi-LoRA Inference Optimization:}  Our method is especially beneficial in multi-LoRA inference settings, where multiple task-specific adapters are deployed. By reducing adapter size without sacrificing accuracy, we significantly enhance inference throughput—crucial for production systems requiring high efficiency and task parallelism.

\textbf{Broad Compatibility with LoRA/PEFT Methods:} Although we show our method's impact with LoRA, our algorithm is compatible with the other parameter-efficient methods, such as DoRA  \cite{liu2024dora}, Pissa \cite{meng2024pissa}, oLoRA \cite{buyukakyuz2024olora}.



\section{Conclusion}


We present LangVision-LoRA-NAS, a novel framework that combines Neural Architecture Search (NAS) with LoRA to optimize Vision-Language Models. Traditional LoRA fine-tuning uses a fixed rank across all layers, which could be suboptimal for computational efficiency. Our method addresses this by dynamically searching for optimal LoRA ranks for each layer, balancing performance. We use weight-sharing NAS to efficiently identify the optimal rank within a predefined search space, reducing the number of trainable parameters while almost maintaining model accuracy. Experiments on the LLaMA-3.2-11B-Vision-Instruct model demonstrate reduction in LoRA parameter counts across diverse open-source downstream datasets, with negligible loss in perplexity. 
For example, the baseline uniform LoRA model uses 268.7M parameters (2.5\% of the total model size), whereas our searched LoRA configuration reduces this to just 103.3M parameters (1.0\% of the model), achieving a 2.6× compression without compromising performance.

\vspace{-0.8ex}



\section*{Acknowledgements}

This research used resources of the Argonne Leadership Computing Facility, a U.S. Department of Energy (DOE) Office of Science user facility at Argonne National Laboratory and is based on research supported by the U.S. DOE Office of Science-Advanced Scientific Computing Research Program, under Contract No. DE-AC02-06CH11357.



\bibliography{main}

\begin{thebibliography}{10}

\bibitem{zhang2024vision}
Jingyi Zhang et~al.,
\newblock ``Vision-language models for vision tasks: A survey,''
\newblock {\em IEEE Transactions on Pattern Analysis and Machine Intelligence}, 2024.

\bibitem{liu2024visual}
Haotian Liu et~al.,
\newblock ``Visual instruction tuning,''
\newblock {\em Advances in neural information processing systems}, vol. 36, 2024.

\bibitem{agrawal2024pixtral}
Pravesh Agrawal et~al.,
\newblock ``Pixtral 12b,''
\newblock {\em arXiv preprint arXiv:2410.07073}, 2024.

\bibitem{beyer2024paligemma}
Lucas Beyer et~al.,
\newblock ``Paligemma: A versatile 3b vlm for transfer,''
\newblock {\em arXiv preprint arXiv:2407.07726}, 2024.

\bibitem{bai2023qwen}
Jinze Bai et~al.,
\newblock ``Qwen-vl: A versatile vision-language model for understanding, localization, text reading, and beyond,''
\newblock {\em arXiv preprint arXiv:2308.12966}, vol. 1, no. 2, pp. 3, 2023.

\bibitem{deitke2024molmo}
Matt Deitke et~al.,
\newblock ``Molmo and pixmo: Open weights and open data for state-of-the-art multimodal models,''
\newblock {\em arXiv preprint arXiv:2409.17146}, 2024.

\bibitem{radford2021learning}
Alec Radford et~al.,
\newblock ``Learning transferable visual models from natural language supervision,''
\newblock in {\em International conference on machine learning}. PMLR, 2021, pp. 8748--8763.

\bibitem{chitty2022neural}
Krishna~Teja Chitty-Venkata et~al.,
\newblock ``Neural architecture search for transformers: A survey,''
\newblock {\em IEEE Access}, vol. 10, pp. 108374--108412, 2022.

\bibitem{vaswani2017attention}
A~Vaswani,
\newblock ``Attention is all you need,''
\newblock {\em Advances in Neural Information Processing Systems}, 2017.

\bibitem{Llama_11B_Vision_Instruct}
meta llama,
\newblock ``Llama-3.2-11b-vision-instruct,'' 2024.

\bibitem{laurençon2024matters}
Hugo Laurençon et~al.,
\newblock ``What matters when building vision-language models?,'' 2024.

\bibitem{kembhavi2016diagram}
Aniruddha Kembhavi et~al.,
\newblock ``A diagram is worth a dozen images,''
\newblock in {\em Computer Vision--ECCV 2016: 14th European Conference, Amsterdam, The Netherlands, October 11--14, 2016, Proceedings, Part IV 14}. Springer, 2016, pp. 235--251.

\bibitem{masry2022chartqa}
Ahmed Masry et~al.,
\newblock ``Chartqa: A benchmark for question answering about charts with visual and logical reasoning,''
\newblock {\em arXiv preprint arXiv:2203.10244}, 2022.

\bibitem{mathew2021docvqa}
Minesh Mathew et~al.,
\newblock ``Docvqa: A dataset for vqa on document images,''
\newblock in {\em Proceedings of the IEEE/CVF winter conference on applications of computer vision}, 2021, pp. 2200--2209.

\bibitem{mathew2022infographicvqa}
Minesh Mathew et~al.,
\newblock ``Infographicvqa,''
\newblock in {\em Proceedings of the IEEE/CVF Winter Conference on Applications of Computer Vision}, 2022, pp. 1697--1706.

\bibitem{lu2021inter}
Pan Lu et~al.,
\newblock ``Inter-gps: Interpretable geometry problem solving with formal language and symbolic reasoning,''
\newblock {\em arXiv preprint arXiv:2105.04165}, 2021.

\bibitem{kembhavi2017you}
Aniruddha Kembhavi et~al.,
\newblock ``Are you smarter than a sixth grader? textbook question answering for multimodal machine comprehension,''
\newblock in {\em Proceedings of the IEEE Conference on Computer Vision and Pattern recognition}, 2017, pp. 4999--5007.

\bibitem{liu2023visual}
Fangyu Liu et~al.,
\newblock ``Visual spatial reasoning,''
\newblock {\em Transactions of the Association for Computational Linguistics}, vol. 11, pp. 635--651, 2023.

\bibitem{lau2018dataset}
Jason~J Lau et~al.,
\newblock ``A dataset of clinically generated visual questions and answers about radiology images,''
\newblock {\em Scientific data}, vol. 5, no. 1, pp. 1--10, 2018.

\bibitem{cheng2021hitab}
Zhoujun Cheng et~al.,
\newblock ``Hitab: A hierarchical table dataset for question answering and natural language generation,''
\newblock {\em arXiv preprint arXiv:2108.06712}, 2021.

\bibitem{kazemi2023geomverse}
Mehran Kazemi et~al.,
\newblock ``Geomverse: A systematic evaluation of large models for geometric reasoning,''
\newblock {\em arXiv preprint arXiv:2312.12241}, 2023.

\bibitem{liu2018darts}
Hanxiao Liu et~al.,
\newblock ``Darts: Differentiable architecture search,''
\newblock {\em arXiv preprint arXiv:1806.09055}, 2018.

\bibitem{stamoulis2019single}
Dimitrios Stamoulis et~al.,
\newblock ``Single-path nas: Designing hardware-efficient convnets in less than 4 hours,''
\newblock in {\em Joint European Conference on Machine Learning and Knowledge Discovery in Databases}. Springer, 2019, pp. 481--497.

\bibitem{liu2024dora}
Shih-Yang Liu et~al.,
\newblock ``Dora: Weight-decomposed low-rank adaptation,''
\newblock {\em arXiv preprint arXiv:2402.09353}, 2024.

\bibitem{meng2024pissa}
Fanxu Meng et~al.,
\newblock ``Pissa: Principal singular values and singular vectors adaptation of large language models,''
\newblock {\em arXiv preprint arXiv:2404.02948}, 2024.

\bibitem{buyukakyuz2024olora}
Kerim B{\"u}y{\"u}kaky{\"u}z,
\newblock ``Olora: Orthonormal low-rank adaptation of large language models,''
\newblock {\em arXiv preprint arXiv:2406.01775}, 2024.

\end{thebibliography}
\bibliographystyle{IEEEbib}
\end{document}